\documentclass[10pt,twocolumn,letterpaper]{article}

\usepackage{array,multirow}
\usepackage{cvpr}
\usepackage{times}
\usepackage{epsfig}
\usepackage{graphicx}
\usepackage{wrapfig}
\usepackage{amsmath}
\usepackage{amssymb}
\usepackage{xcolor}
\usepackage{caption}
\usepackage[utf8]{inputenc}

\usepackage{mathtools}
\usepackage{algorithm}
\usepackage[noend]{algpseudocode}

\usepackage[pagebackref=true,breaklinks=true,letterpaper=true,colorlinks,bookmarks=false]{hyperref}

\cvprfinalcopy 

\def\cvprPaperID{****} 
\algrenewcommand\algorithmicindent{1.1em}%

\ifcvprfinal\fi

\begin{document}

\title{An Autoencoder-based Learned Image Compressor:\\Description of Challenge Proposal by NCTU}

\author{
    \begin{tabular}{cccc}
        David Alexandre &
        Chih-Peng Chang &
        Wen-Hsiao Peng &
        Hsueh-Ming Hang \\
        {\small davidalexandre.eed05g@nctu.edu.tw} &
        {\small gcwhiteshadow.cs04@nctu.edu.tw} &
        {\small wpeng@cs.nctu.edu.tw} &
        {\small hmhang@nctu.edu.tw} 
    \end{tabular}\\[3ex]
    National Chiao Tung University, Taiwan
    \vspace{-1ex}
}

\maketitle
\thispagestyle{empty}

\begin{abstract}
   We propose a lossy image compression system using the deep-learning autoencoder structure to participate in the Challenge on Learned Image Compression (CLIC) 2018. Our autoencoder uses the residual blocks with skip connections to reduce the correlation among image pixels and condense the input image into a set of feature maps, a compact representation of the original image. The bit allocation and bitrate control are implemented by using the importance maps and quantizer. The importance maps are generated by a separate neural net in the encoder. The autoencoder and the importance net are trained jointly based on minimizing a weighted sum of mean squared error, MS-SSIM, and a rate estimate. Our aim is to produce reconstructed images with good subjective quality subject to the 0.15 bits-per-pixel constraint.
\end{abstract}

\section{Introduction}

Lossy image compression has been a challenging topic in deep learning domain in recent years. Until now, several methods have been proposed for compressing images using deep learning techniques, for example, end-to-end trained neural networks, post-processing neural networks with non-neural codecs, or improving traditional codecs using neural networks. The autoencoder structure has been a popular choice to do end-to-end image compression. The encoder part learns to generate a compact representation of the input image, and the decoder part reconstructs an image close to the input based on the compact representation. Similar to the traditional techniques, the performance of a lossy image compression scheme relies on how well the quantization and rate control scheme are blended into this autoencoder structure. 

Our response to this Challenge on Learned Image Compression (CLIC) at CVPR 2018 is an end-to-end trainable autoencoder with residual blocks and importance maps. More specifically, our system aims to meet the 0.15 bits-per-pixel requirement while achieving good Peak Signal Noise Ratio (PSNR) performance and subjective quality. Our neural net architecture is inspired by the works of \cite{ETHZ} and \cite{imp_map}. We adopt the autoencoder architecture from Mentzer \emph{et al.} \cite{ETHZ} and the importance map concept from Li \emph{et al.} \cite{imp_map}. The architecture from Mentzer \emph{et al.} consists of a set of symmetrical convolution layers and residual block layers to compress and decompress an image. The encoder reduces the dimensionality of an image into a number of feature maps. The feature maps contain densely packed color values from an image and can be reconstructed closely into its original value by using the associated decoder. Each feature map is further modulated with an importance map, which controls the bit allocation among feature samples in a feature map, determining critically the final compression rate of the output image. The importance maps are generated by a set of trained convolution layers that can identify the important feature samples (i.e., samples that need more bits) of an image and produce a set of importance masks to truncate the (bit) depth of feature maps in a spatially varying manner. Therefore, the final feature maps keep only the most meaningful information of representing an image at a specific rate. They are essentially the compact representation of the input image.

To overcome the training problem due to the quantization operation, we adopt the soft quantization technique in  \cite{end_to_end} and implement a straightforward rate estimation based on the values of importance masks for the rate control purpose. Our entropy coder, which compresses the truncated feature maps, is developed based on the JPEG2000 standard. We adjust the context model (in the arithmetic coder) to match our feature map characteristics, which contains all zero at the least significant bitplane in most of the cases.

\begin{figure}
\includegraphics[width=\linewidth]{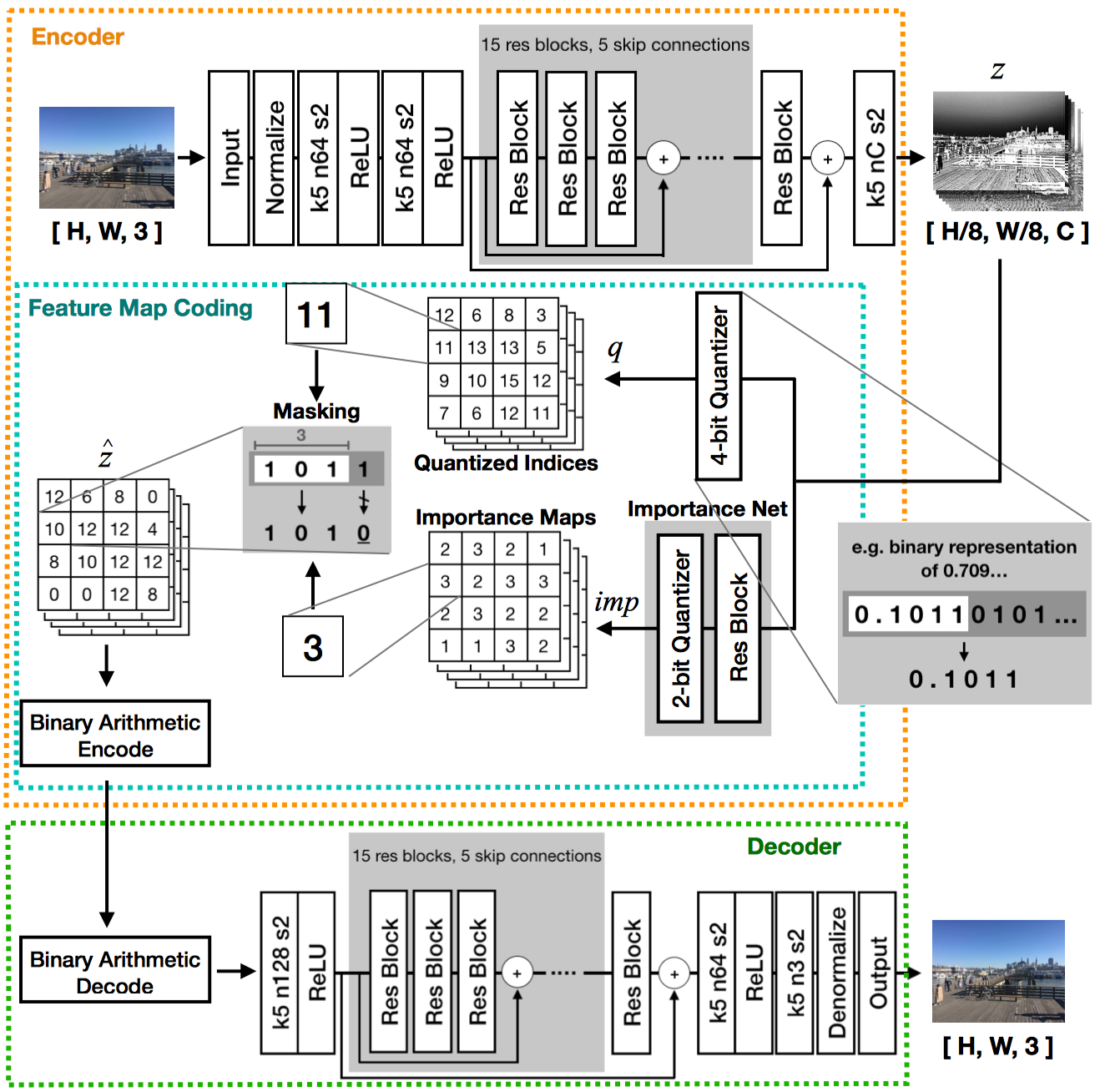}
\caption{The proposed network architecture.}
\label{fig:architecture}
\vspace{-0.25cm}
\end{figure}

As instructed by the challenge guidelines, PSNR is used to measure the output image quality. We compare it with the popular image compression standards such as JPEG and BPG. Also, we use MS-SSIM, which mimics human subjective quality evaluation, as the second image quality metric.

\section{Methods}

We give more details of our implementation in this section, which includes the deep learning network architecture, the generation of importance map, and finally, the quantization and arithmetic coder in our system.

\subsection{Network Architecture}

The autoencoder part for feature map generation and image reconstruction is adapted from \cite{ETHZ}. The input to the encoder is an image with dimension H$\times$W in RGB format. The upper part represents the encoder $E$ and the lower part represents the decoder $D$. k5 n64 s2 represents kernel size 5, with 64 channels and stride 2. All layers in the encoder are the linear convolution and non-linear activation operations, and the decoder layers implement the inverse operations.

Given an input image $x$, the encoder produces a set of feature maps $z = E(x)$. The normalization at the encoder limits the image input value to $[-1, 1]$. The encoder output $z$, limited in the range of 0 to 1, goes to the quantizer and the importance net. The quantizer, $Q$, produces a fixed-point representation $q = Q(z)$ of $z$  by keeping the first $d$ bits after the decimal point. Essentially, this amounts to performing a uniform quantization on $z$ with a step size of $1/2^d$. In the meanwhile, $z$ is the input to the \textit{importance net}, $I$, and it generates an importance map, $imp = I(z)$, for each feature map. In particular, the values of the importance maps are positive integers ranging from 1 to $d$, indicating how many (most significant) bits of the quantized feature $q$ should be kept via the masking operation  $\hat{z} = q \cdot imp$. The decoder, $D$, further takes in $\hat{z}$ after the binary arithmetic decoding of the bitstream to generate the reconstructed image $\hat{x} = D(\hat{z})$. The denormalization at the end of decoder reverses the encoder input normalization, clipping the output pixel value to $[0, 255]$. 

\subsection{Quantizer}

The quantizer is a critical element in lossy image compression. Our quantizer is located at the end of encoder. Specifically, the quantized value  of a feature sample is obtained as follows:
\begin{equation}
\label{eq:hard_q}
q =Q(z)=\ \left \lfloor z \times 2^{d} \right \rfloor,
\end{equation}
where $d$ is the desired bit depth (which is 4 in our submitted codec), $\left\lfloor \cdot \right\rfloor$ denotes the floor function, and $z = E(x)$ is the feature sample. In the training phase, we use the following soft-quantization \cite{end_to_end} in place of the hard quantization in \eqref{eq:hard_q} to facilitate the gradient calculation in back-propagation:
\begin{equation}
\tilde{z} = \sum_{i=0}^{2^{d}-1}\frac{\exp(- \| z \times 2^d-i \|)}{\sum_{j=0}^{2^d-1} \exp (- \| z \times 2^d-j \|)} \times i.
\end{equation}

Essentially, this soft-quantization approximates the quantizer mapping function of the hard-quantization with a smooth differentiable function so as to have non-zero gradients at training time. In addition, we save the quantized values with a series of integer indices to save space when storing the multiple quantized values. These indices can be represented in binary form to match the masking operation described in the next subsection.

\subsection{Importance Net}

\begin{figure}
\includegraphics[width=\linewidth]{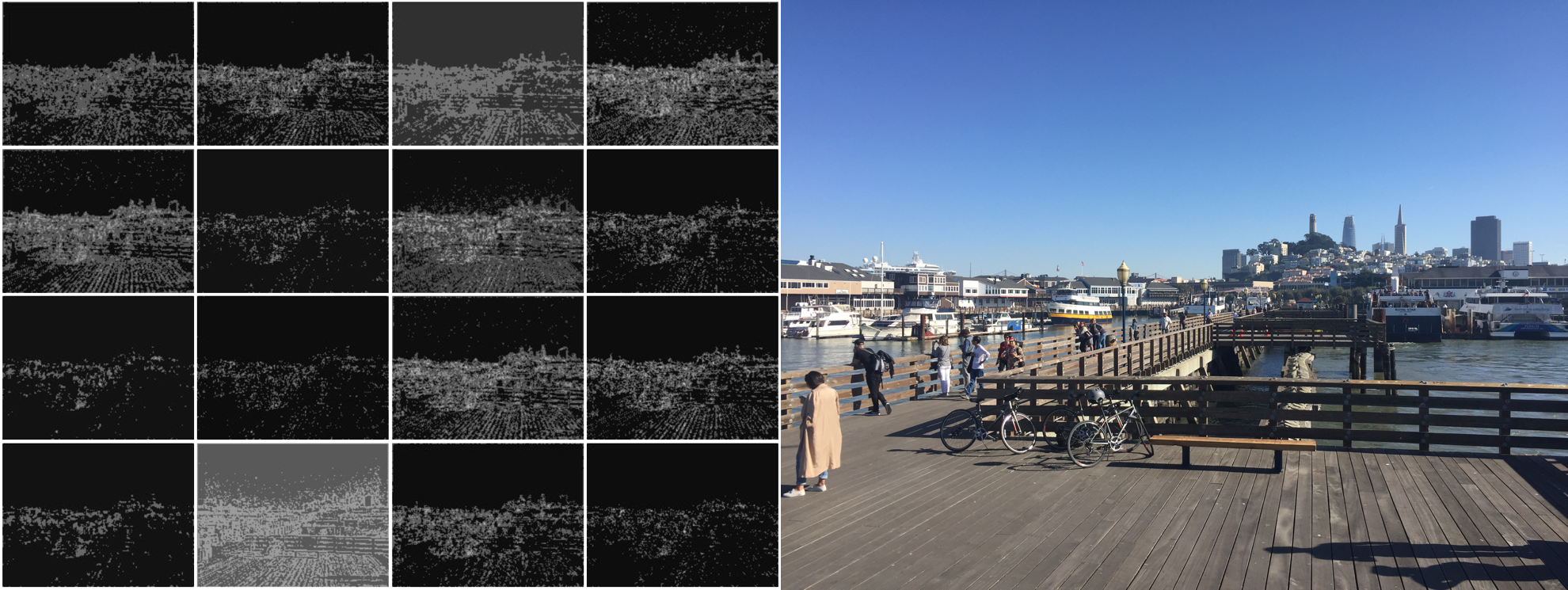}
\caption{Importance maps (left) for an input image (right).}
\label{fig:imps}
\vspace{-0.25cm}
\end{figure}

The importance map addresses the issue of content complexity variation in different parts of an image. Also, some images are easier to compress than others. It is closely related to measuring the complexity of an image. Li \emph{et al.} \cite{imp_map} introduce the importance map as the solution to this problem. It is expected that the importance map, produced by a neural network, can learn to signify the complexity of different areas in an image. Thus, at a given bitrate, the compression system allocates more bits to the complicated areas.

Our importance net is made of residual blocks, followed by a quantizer to produce an integer-valued importance map $imp$ for each feature map. The $imp$ acts as a mask to set zero bits of the quantized feature $q$ which are less significant in contributing to the overall reconstruction quality, as done by \eqref{eq:masking} and illustrated in Figure \ref{fig:imps}. In other words, for each sample on the feature map, the corresponding $imp$ value, $m$, determines how many bits of information to retain on the final feature map. As such, the importance maps have a resolution and  a channel number identical to the feature maps. It is worth noting that they are produced based on the same encoder output $z$. In the training process, the quantizer in the importance net is replaced by the soft-quantization in \cite{end_to_end}. Figure \ref{fig:imps} shows the importance maps for a sample image.
\begin{equation}
\label{eq:masking}
\hat{z}= \left \lfloor \frac{q}{2^{d-m}} \right \rfloor \times 2^{d-m}
\end{equation}

\subsection{Loss Function}

Our loss function is a weighted sum of rate and distortion, $\lambda R + D$, as shown in \eqref{eq:DRloss}. The parameter $\lambda$ controls the trade-off between the rate loss and the distortion loss. We estimate the rate loss by summing up all values of the importance maps, $H(imp)$. For distortion, we calculate both the Mean Squared Error (MSE) and the Multi-Scale Structural Similarity (MS-SSIM) index between the original image $x$ and the reconstructed image $\hat{x}$ as in \eqref{eq:Dloss}, where both $\sigma_{1}$ and $\sigma_{2}$ are trained to minimize \eqref{eq:DRloss}. 
\begin{equation}
\label{eq:DRloss}
L = \lambda \times H( imp) \ +\ L_{D}
\end{equation}
\begin{equation}
\label{eq:Dloss}
L_{D} = \frac{MSE}{2\sigma ^{2}_{1}} + \frac{MS-SSIM}{2\sigma ^{2}_{2}}  +\log \sigma ^{2}_{1} +\log \sigma ^{2}_{2}
\end{equation}

\subsection{Entropy Coding}

\begin{figure}
\centering
\includegraphics[width=0.8\linewidth]{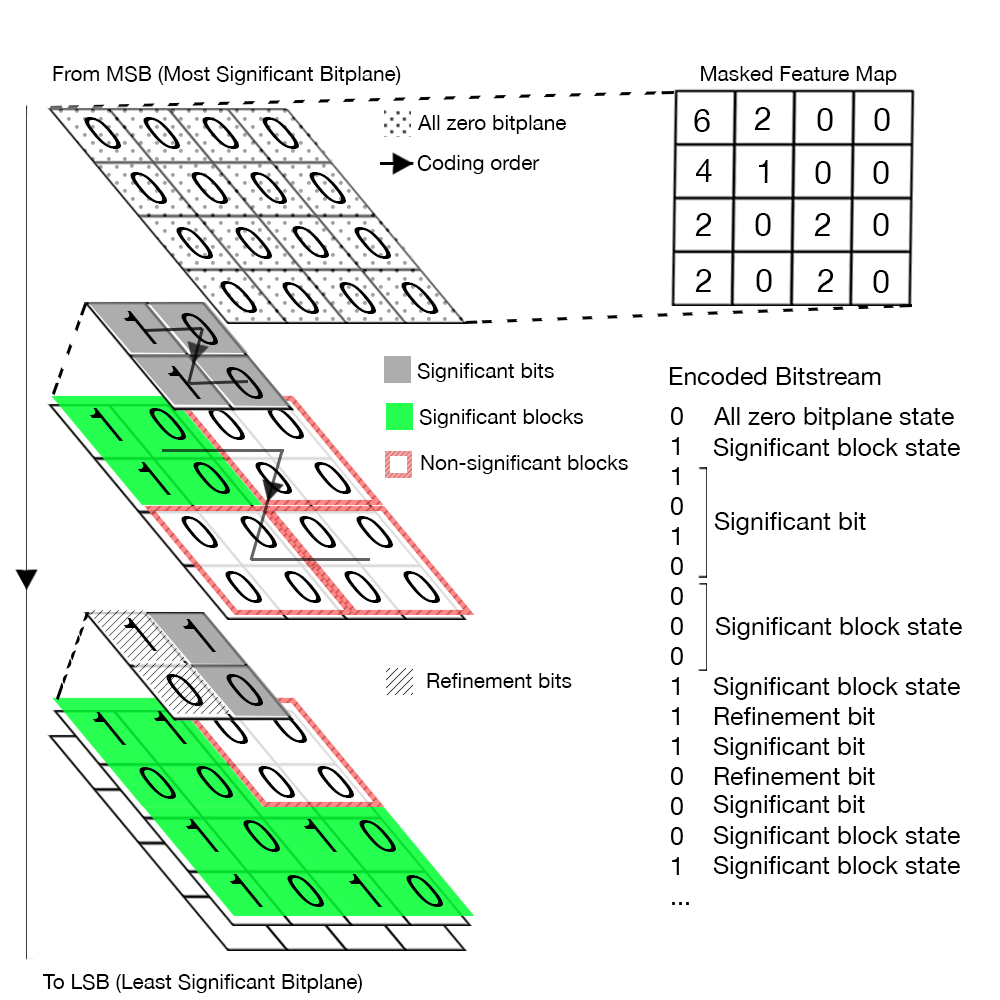}
\vspace{-0.25cm}
\caption{Binary arithmetic coding of feature maps.}
\label{fig:bac}
\vspace{-0.35cm}
\end{figure}

\begin{algorithm}[t]
\caption{Binary arithmetic coding of feature maps}\label{euclid}
\begin{algorithmic}[1]
\State Initialize final feature maps Z
\For{each feature maps}
\State Initialize Block Flag $B_{m, n}$ to 0
\State Initialize Sample Flag $S_{m, n, i, j}$ to 0
\State Initialize All Zero Bitplane Flag $AZB$ to 1
\For{each bitplane $k$ = MSB to LSB}
\If{$AZB == 1$}
\State Set $AZB$ as per all zero bitplane state
\State Encode $AZB$ 
\EndIf
\If{$AZB == 0$}
\For{each block $m, n$}
\If{$B_{m, n} == 0$}
\State Set $B_{m, n}$ as per significant block state
\State Encode $B_{m, n}$ 
\EndIf
\If{$B_{m, n} == 1$}
\For{each sample $i, j$}
\If{$S_{m, n, i, j} == 0$}
\State Set $S_{m, n, i, j}$ as per significance state 
\State Encode $significant~bit$
\Else
\State Encode $refinement~bit$
\EndIf
\EndFor
\EndIf
\EndFor
\EndIf
\EndFor
\EndFor
\end{algorithmic}
\end{algorithm}

Our entropy coding is inspired by the JPEG2000 standard. It encodes the binary representation of individual final feature maps  from the most significant bitplane (MSB) to the least significant one (LSB). Furthermore, on each bitplane, the coding bits are classified into the significant and refinement types, and coded on a block by block basis through a context-adaptive binary arithmetic coder. The parameter for block size is experimental. As with JPEG2000, separate context models are used for different types of bit, with the context probabilities updated constantly along the way. Additionally, we introduce flags at both the bitplane and block levels to indicate the all zero cases. More details are presented in Algorithm 1. 

The entropy decoder decodes the compressed bitstream and recover the binary feature maps. The encoded bitstream also contains side information needed by the decoder. Figure \ref{fig:bac} illustrates the procedure of our entropy coding.

\section{Training}

This section describes our training procedure and parameter settings.

\subsection{Datasets}

Our training dataset contains 10,000 images from the ImageNet Large Scale Visual Recognition Competition (ILSVRC) 2012 \cite{ILSVRC2015}. For training, we randomly crop  128x128 patches from these images as inputs to the autoencoder. For testing and validation, we use Kodak PhotoCD dataset. 

\subsection{Procedure}

The training for the autoencoder and importance net is done in three sequential phases. In the first pre-training phase, the autoencoder is trained independently without the importance net. This is achieved by minimizing the reconstruction distortion alone. In the second phase, the importance net is included in the training process, to minimize an objective function that involves both the distortion and rate estimate. In this phase, however, the pre-trained autoencoder is not updated. In the final fine-tuning phase, both the autoencoder and the importance net are updated jointly. Table \ref{tab:ps} shows the detailed parameter settings.

\begin{table}
\begin{center}
\begin{tabular}{l  r}
Parameters              & Values                \\
\hline
$\lambda$                       & [0.001, 10]   \\
Batch size                      & 10                    \\
Bit depth ($d$)                       & 4                     \\
Block size                      & 4 $\times$ 4  \\
Learning rate           & [1e-4, 1e-6]  \\
Channel number      & 16                    \\
\hline
\end{tabular}
\end{center}
\vspace{-0.5cm}
\caption{Parameter settings.}
\label{tab:ps}
\vspace{-0.2cm}
\end{table}

\section{Evaluation}

This section compares the compression performance of our system with that of JPEG and BPG at ~0.15 bpp in terms of PSNR and MS-SSIM. The test dataset is from CLIC 2018. The results are presented in Table \ref{tab:CPC}. 

\begin{table}
\begin{center}
\begin{tabular}{l  c c c}
& BITS/PIXEL           & PSNR  & MS-SSIM                       \\
\hline
JPEG            & 0.155 & 22.1  & 0.7568                \\
BPG             & 0.153 & 29.2  & 0.9425                \\
Ours            & 0.126 & 26.3  & 0.9242
\\
\hline
\end{tabular}
\end{center}
\vspace{-0.5cm}
\caption{Compression performance comparison.}
\label{tab:CPC}
\vspace{-0.3cm}
\end{table}

\begin{figure}
\begin{tabular}{cc}
\includegraphics[width=0.45\linewidth]{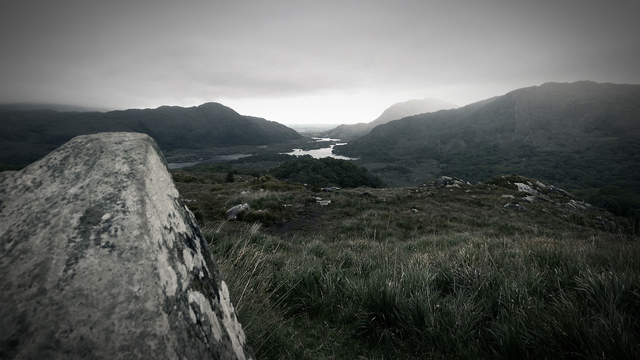}
& \includegraphics[width=0.45\linewidth]{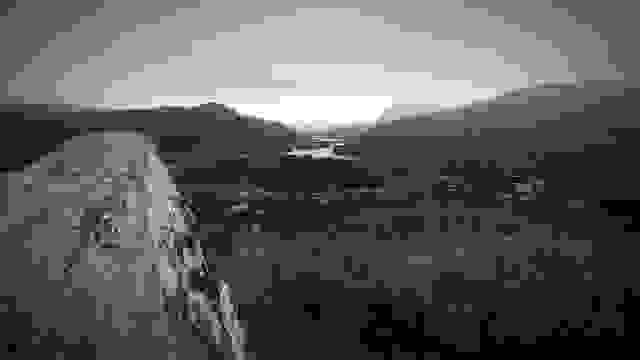}\\
(a) Original image       &       (b) JPEG\\
\includegraphics[width=0.45\linewidth]{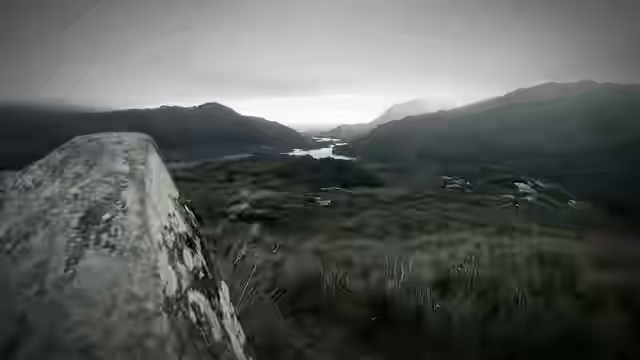}
& \includegraphics[width=0.45\linewidth]{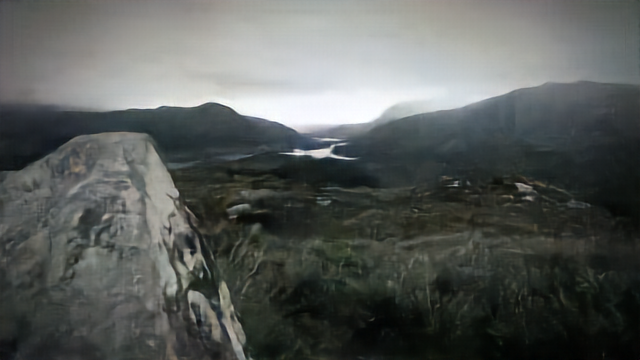}\\
(c) BPG  &       (d) Our System
\end{tabular}
\vspace{-0.2cm}
\caption{Original image (a) and the reconstructed images produced by (b) JPEG , (c) BPG, and (d) our system.}
\label{fig:comparison}
\vspace{-0.4cm}
\end{figure}

It is worth noting that the PSNR numbers are computed following the CLIC 2018 method; that is, the mean squared error over these 287 test images is first calculated before the PSNR conversion. For MS-SSIM, we acquire the numbers by averaging the MS-SSIM values of these test images, with the MS-SSIM of each test image obtained by taking the average over the three color components. 
From Table \ref{tab:CPC}, both BPG and our system outperform JPEG by a large margin in terms of PSNR. Although the BPG is 3dB better than ours in terms of PSNR, their MS-SSIM are close to each other. Figure \ref{fig:comparison} presents few sample images to compare the subjective quality of these methods.

\section{Conclusion}

This work describes NCTU’s response to the CLIC 2018. Its presents an autoencoder-based learned image compressor with the notion of importance maps for bit allocation and rate control. While it outperforms JPEG significantly, there is still a performance gap relative to BPG, suggesting ample room for further improvements. Among others, the impact of importance maps on subjective quality deserves further investigation. In our case, it causes color banding in several images.

\section{Acknowledgement}

This work was supported in part by the MOST, Taiwan under Grant MOST 107-2634-F-009-007. The authors are grateful to the National Center for High Performance Computing, Taiwan, for providing GPU facility. Also, we like to thank Ms. Yang-Tzu Liu Tsen for her help on the computer simulation and arithmetic coder implementation.

\renewcommand\refname{Reference}
\bibliographystyle{unsrt}

\end{document}